\documentclass[runningheads]{llncs}

\usepackage{eccv}

\usepackage{eccvabbrv}

\usepackage{graphicx}
\usepackage{booktabs}
\usepackage{amsmath,amssymb,bm}
\usepackage{mathtools}
\usepackage{algorithm}
\usepackage{algpseudocode}
\usepackage{adjustbox}
\DeclareMathOperator*{\argmin}{argmin}
\DeclareMathOperator*{\argmax}{argmax}

\usepackage[accsupp]{axessibility}  % Improves PDF readability for those with disabilities.

\usepackage[pagebackref,breaklinks,colorlinks,citecolor=eccvblue]{hyperref}
\usepackage{orcidlink}

\begin{document}

\title{High-Frequency Anti-DreamBooth: Robust Defense against Personalized Image Synthesis}

\titlerunning{High-Frequency Anti-DreamBooth: Robust Defense Against Image Synthesis}

\author{Takuto Onikubo\inst{} \and
Yusuke Matsui\inst{}\orcidlink{0000-0003-1529-0154}}

\authorrunning{T.~Onikubo et al.}

\institute{The University of Tokyo}

\maketitle

\begin{abstract}
    Recently, text-to-image generative models have been misused to create unauthorized malicious images of individuals, posing a growing social problem.
    Previous solutions, such as Anti-DreamBooth, add adversarial noise to images to protect them from being used as training data for malicious generation.
    However, we found that the adversarial noise can be removed by adversarial purification methods such as DiffPure.
    Therefore, we propose a new adversarial attack method that adds strong perturbation on the high-frequency areas of images to make it more robust to adversarial purification.
    Our experiment showed that the adversarial images retained noise even after adversarial purification, hindering malicious image generation.

\end{abstract}

\section{Introduction}
    \label{sec:intro}
    Recently, the progress of text-to-image generative models has been remarkable.
    One of the practical applications of these models is personalization, a fine-tuning method to generate images of a specific subject with a few examples.
    However, some people personalize models to generate unauthorized malicious images called fake images, posing a significant social problem.
    For example, some people download photos of celebrities from the internet and fine-tune generative models with these photos to generate scandalous images~\cite{fakenews}.
    Furthermore, some painters are losing their jobs due to the generation of artworks that mimic their techniques~\cite{illustrating}.
    
    To address the issue of fake images, Thanh et al.~\cite{le_etal2023antidreambooth} proposed an adversarial attack named Anti-DreamBooth, a method that prevents personalized generation.
    We can use Anti-DreamBooth to add slight noise perturbation to our images before publishing them, hindering any unauthorized personalized generations.
    Additionally, Shan et al.~\cite{shan2023glaze} proposed Glaze to prevent the generation of images that mimic paintings created by artists.
    Although these methods are very powerful, we have found that the added perturbations can be removed easily by adversarial purification techniques such as DiffPure~\cite{nie2022diffusion}.
    We also found they are vulnerable to simple noise-removal filters like Gaussian or bilateral filters.

    To address the problem, we propose an adversarial attack that adds strong adversarial noise on high-frequency areas of images.
    Our method utilizes a high-pass filter to create a mask that represents the edges of the image and adds especially strong adversarial noise to the masked area.
    The intense adversarial noise in the complicated region is difficult to remove despite being inconspicuous.
    
\begin{figure*}[t]
        \centering
        \includegraphics[width=11cm]{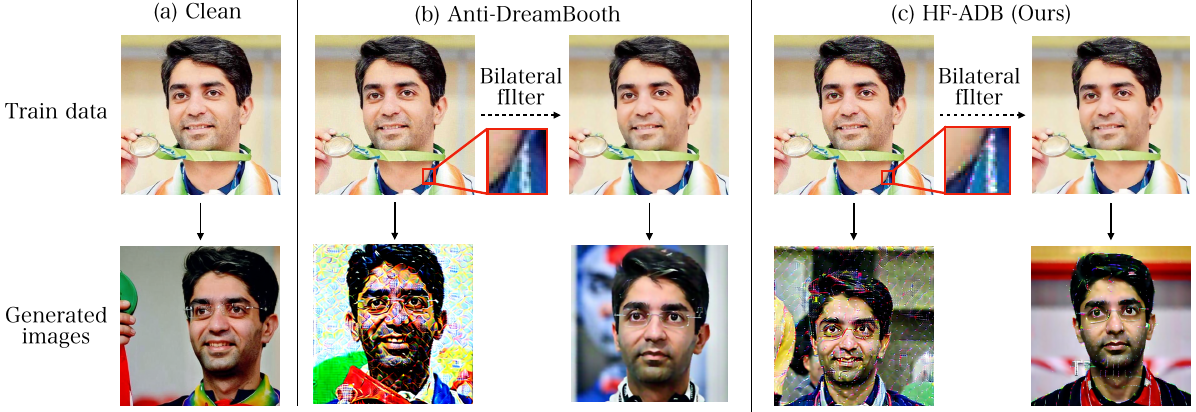}
        \caption{Summary: (a) We can generate realistic images by personalizing a model. (b) Applying Anti-DreamBooth~\cite{le_etal2023antidreambooth} to the train images hinders personalized generation. However, we can break the defense by noise removal methods such as bilateral filters. (c) Our method maintains its defense even after applying noise removal methods.}
        \label{fig:overall_result}
    \end{figure*}

    The overview of our work is shown in Fig. \ref{fig:overall_result} with the following contributions.
    \begin{itemize}
        \item We found that the adversarial noise previous works add can be removed easily by adversarial purification technique.
        \item By adding strong perturbation to the high-frequency areas on images, we propose a robust adversarial attack against adversarial purification.
        \item As our idea to vary the strengths of noise does not depend on a specific adversarial attack, we can apply the concept to various adversarial attacks.
    \end{itemize}
    
\section{Related Work}
    \subsection{Adversarial Attack and Adversarial Purification}
    Adversarial attack~\cite{42503, goodfellow2014explaining,chakraborty2018adversarial} is a method that causes a model to fail by taking advantage of its learning and operation processes.
    For example, there are attacks against speech recognition~\cite{carlini2016hidden,zhang2017dolphinattack} or image recognition~\cite{kong2020physgan,wu2023adversarial}.
    Also, data poisoning~\cite{rubinstein2009antidote,yang2017generative} is an adversarial attack that prevents the model from being trained adequately by injecting malicious data into the dataset.
    
    % \subsection{Adversarial Purification}
    As a countermeasure against adversarial attacks, a method called adversarial purification~\cite{samangouei2018defense} uses generative models to purify the adversarial noise.
    DiffPure~\cite{nie2022diffusion} is an adversarial purification method that uses diffusion models. 
    Although computationally expensive, DiffPure is one of the most potent adversarial purification methods.
    
    \subsection{Abuse of Text-to-Image Generative Models}
    Generating fake images or imitating illustrations by personalizing generative models has become a significant problem.
    In order to handle the issue, adversarial attack-based methods~\cite{le_etal2023antidreambooth,shan2023glaze} and watermarking-based methods~\cite{liu2024countering,zhu2024watermark} have been mainly studied.

\section{Preliminaries}
    In this section, we describe the problem setting.
    Suppose a user uploads their images onto the internet.
    Malicious people might collect the images, personalize pre-trained models, and generate manipulated images (e.g., depicting the user committing a crime).
    To prevent personalization, we suggest a method for users to protect their images, i.e., adding adversarial perturbations to their images before uploading them to the internet.

    We can formulate the problem as follows.
    First, we denote $\left\{\mathbf{x}^{(\textit{i})}\right\}_{i=1}^{N_{\rm{db}}}\subset[0,~1]^{H\times W\times3}$ as the images for the user to protect.
    Here, $N_{\rm{db}} \in \mathbb{N}$ is the number of images to upload to the internet.
    Next, we optimize perturbations $\left\{\boldsymbol{\delta}^{(\textit{i})}\right\}_{i=1}^{N_{\rm{db}}}\subset{[0,~1]^{H\times W\times 3}}$.
    We then construct adversarial examples by adding the perturbations to the images: $\left\{\mathbf{x}^{(\textit{i})}+\boldsymbol{\delta}^{(\textit{i})}\right\}_{i=1}^{N_{\rm{db}}}$.
    Our task is to maximize the output of the loss function $\mathcal{L_{\rm{db}}}$ during fine-tuning using these adversarial examples so that the fine-tuned model fails to generate the user's images.

\section{Method}
In this section, we propose a method to hinder personalized image generation using DreamBooth~\cite{ruiz2023dreambooth} by adding strong adversarial perturbation to the high-frequency areas of the images.
Our method adds perturbation that is difficult to remove using traditional filters or adversarial purification techniques.

\begin{figure*}[t]
    \begin{center}
        \includegraphics[width=1\linewidth]{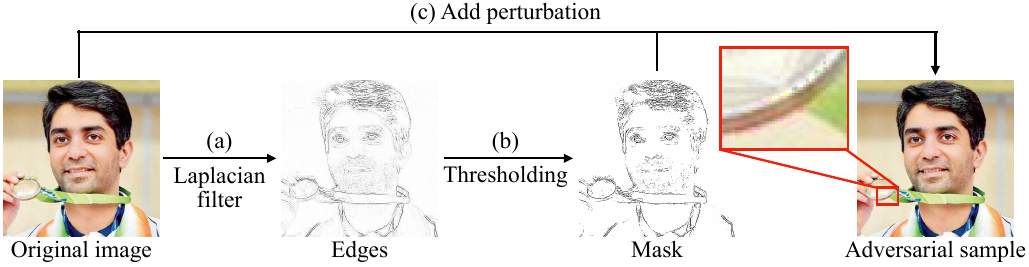}
        \caption{Procedure steps of our method.}
        \label{fig:method}
    \end{center}
\end{figure*}

Fig. \ref{fig:method} illustrates the workflow of our method.
First, we apply a $3\times3$ Laplacian filter to the input image and extract the edges of the image (Fig. \ref{fig:method}(a)).
Then, by thresholding the pixel values of the edges, we obtain a mask of the edges (Fig. \ref{fig:method}(b)).
This mask shows the area with high frequency.
The threshold is determined for each image so that the ratio of the masked area is constant (in practice, 3-5\% of the entire image).
Next, we add adversarial perturbation using the created mask (Fig. \ref{fig:method}(c)).
As shown in the red box area, strong perturbation is added only to the edges, while weak perturbation is added to the entire image.

Here, we used ASPL (Alternating Surrogate and Perturbation Learning) from Anti-DreamBooth~\cite{le_etal2023antidreambooth} to calculate the noise.
Our goal is to generate adversarial examples that degrade the performance of a model fine-tuned on these images.
This is too complicated to solve.
Therefore, ASPL exploits a model called surrogate model $\boldsymbol{\theta}_{\rm{sur}}$, an approximation of a model personalized with adversarial examples.
ASPL trains a surrogate model using the user's reference images $\mathcal{X}$ and simultaneously calculates adversarial perturbations that degrade the performance of the surrogate model.

We can formulate the process of adding perturbation as Alg.~\ref{alg:method}.
Firstly, we initialize the adversarial noise and the surrogate model parameters in L\ref{alg:init1} and L\ref{alg:init2}.
Then, we add adversarial noise over $T$ steps in L\ref{loop_start}--\ref{loop_end}.
Specifically, in L\ref{db1}, the surrogate model $\boldsymbol{\theta}_{\rm{sur}}$ is fine-tuned for one step, resulting in the model $\boldsymbol{\theta}_{\rm{tmp}}$. Here, $\mathcal{L}_{\rm{db}}$ is the loss function of DreamBooth \cite{ruiz2023dreambooth}.
Next, in L\ref{add_noise}, perturbation is added to each pixel of the image by $\eta_{\rm{unit}}$ so that the inference loss of the model $\boldsymbol{\theta}_{\rm{tmp}}$ is maximized. Here, $\mathcal{L}_{\rm{cond}}$ is the loss function of a typical diffusion model. Then, in L\ref{clamp}, adversarial perturbation is clamped so that the perturbation on the masked area is less than $\eta_{\rm{mask}}$, and otherwise less than $\eta$.
This part is the core of our method.
Through this process, we can create adversarial examples with strong adversarial perturbations at the edges of the image.
Finally, in L\ref{db2}, the surrogate model is fine-tuned for one step using the updated adversarial examples.
We repeat these steps over $T$ steps.

    \begin{algorithm}[t]
        \caption{Algorithm of our method}
        \label{alg:method}
        \begin{algorithmic}[1]
            \Require{
                \begin{tabular}{lll}
                    \multicolumn{2}{l}{$\left\{\mathbf{x}^{(\textit{i})}\right\}_{i=1}^{N_{\rm{db}}} \subset [0,~1]^{H\times W\times 3}$}&\multicolumn{1}{l}{:~original images,} \\ 
                    \multicolumn{2}{l}{$\left\{\mathbf{m}^{(\textit{i})}\right\}_{i=1}^{N_{\rm{db}}} \subset [0,~1]^{H\times W\times 3}$}&\multicolumn{1}{l}{:~masks,} \\
                    \multicolumn{1}{l}{$\mathcal{X}\subset [0,~1]^{H\times W\times 3}$}&\multicolumn{2}{l}{:~reference images for the surrogate model fine-tuning,} \\
                    \multicolumn{1}{l}{$\eta \in [0,~1]$}&\multicolumn{2}{l}{:~maximum magnitude of the weak perturbation,} \\
                    \multicolumn{1}{l}{$\eta_{\rm{mask}} \in [\eta,~1]$}&\multicolumn{2}{l}{:~maximum magnitude of the strong perturbation,}\\
                    \multicolumn{1}{l}{$\eta_{\rm{unit}} \in [0,~\eta]$}&\multicolumn{2}{l}{:~the amount of perturbation added in one step,} \\
                    \multicolumn{1}{l}{$\boldsymbol{\theta}_{\rm{pre}}$}&\multicolumn{2}{l}{:~pre-trained parameter of Stable Diffusion,} \\ 
                    \multicolumn{1}{l}{$T$}&\multicolumn{2}{l}{:~the number of steps to add perturbation,}

                \end{tabular}
            }
            \Ensure{
                $\left\{\boldsymbol{\delta}^{(\textit{i})}_{T}\right\}_{i=1}^{N_{\rm{db}}}\subset{[0,~1]^{H\times W\times 3}}\textrm{:~adversarial perturbation.}$
            }
            \State $\boldsymbol{\delta}_{0}^{(\textit{i})} \gets \mathbf{0} \textrm{~(for all $i$)~:~initialization of noise }\boldsymbol{\delta} $
            \label{alg:init1}
            \State $\boldsymbol{\theta}_{\rm{sur}} \gets \boldsymbol{\theta}_{\rm{pre}} \textrm{~~~~~~~~~~:~initialization of the parameters of the surrogate model}$
            \label{alg:init2}
            \For{$t \in \{1,\dots,T\}$}
                \label{loop_start}
                \State $\boldsymbol{\theta}_{\rm{tmp}} \gets \underset{\boldsymbol{\theta}}{\argmin}\underset{\bf{x_{\rm{ref}}}\in \mathcal{X}}{\sum}\mathcal{L}_{\rm{db}}(\boldsymbol{\theta},~\bf{x}_{\rm{ref}})$~ (with $\boldsymbol{\theta}_{\rm{sur}}$ as the initial weight)
                    \label{db1}
                \State $\boldsymbol{\delta}^{(\textit{i})} \gets \underset{\boldsymbol{\delta} \in [0,~\eta_{\rm{unit}}]^{H\times W\times3}}{\argmax}\ \mathcal{L}_{\rm{cond}}(\boldsymbol{\theta}_{\rm{tmp}},~\mathbf{x}^{(\textit{i})}+\boldsymbol{\delta}_{\textit{t}-1}^{(\textit{i})}+\boldsymbol{\delta})\textrm{~(for all $i$)~}$
                    \label{add_noise}
                \State $\boldsymbol{\delta}_{t}^{(\textit{i})} \gets \rm{clamp}\left(\boldsymbol{\delta}_{\textit{t}-1}^{(\textit{i})}+\boldsymbol{\delta}^{(\textit{i})},~\mathbf{m}^{(\textit{i})},~\eta,~\eta_{\rm{mask}}\right)\textrm{~(for all $i$)~}$
                    \label{clamp}
                \State $\boldsymbol{\theta}_{\rm{sur}} \gets \underset{\boldsymbol{\theta}}{\argmin}\sum_{i=1}^{N_{\rm{db}}}\mathcal{L}_{\rm{db}}(\boldsymbol{\theta},~\mathbf{x}^{(\textit{i})}+\boldsymbol{\delta}_{t}^{(\textit{i})})$
                    \label{db2}
            \EndFor
                \label{loop_end}
        \end{algorithmic}
    \end{algorithm}

\section{Experiment}
    \subsection{Dataset}
        In this study, we use VGGFace2~\cite{cao2018vggface2} as the dataset. VGGFace2 is a large-scale dataset for face recognition, containing over 3.3 million face images of more than 9,000 individuals.
        We sampled 400 face images from 50 individuals from this dataset, with eight images per individual. To eliminate the influence of image quality on the experimental results, we selected images with a resolution of no less than $512\times 512$ pixels and resized them to $512\times 512$ pixels.
        
        We divided eight samples of each individual into two groups of four images.
        The first four images were training images $\left\{\mathbf{x}^{(\textit{i})}\right\}_{i=1}^{N_{\rm{db}}}$ for personalized generation, and the remains were training images $\mathcal{X}$ for the surrogate model.
        We define the former as the original images and the latter as the reference images.

    \subsection{Experiment Settings}
        Firstly, we generated adversarial examples for original images.
        Here, we changed the strength of the adversarial perturbation $\eta, \eta_{\rm{mask}}$ and the ratio of the masked area as the hyper-parameters.
        Then, we purify the added perturbations.
        We used bilateral filter~\cite{paris2009bilateral} and DiffPure~\cite{nie2022diffusion} for purification.
        Next, we personalized Stable Diffusion with these images.
        Finally, we generated images with a prompt of \texttt{"a photo of sks person"} with the personalized model.
        Here, \texttt{"sks person"} is a pseudo-class of target individuals used for personalization.

        In this experiment, we used Stable Diffusion v2.1 as the Text-to-Image generative model and DreamBooth~\cite{ruiz2023dreambooth} as the personalization method.
        Adding adversarial noise to each image took about 2 minutes on an NVIDIA A100 GPU.

        \label{sec:evaluation}
        To evaluate the experimental results, we used the following metrics.
        Firstly, FDSR (Face Detection Success Ratio) is the probability of being successfully detected by RetinaFace~\cite{deng2019retinaface}.
        Then, ISM (Identity Score Matching) measures the similarity of the two images by comparing the ArcFace~\cite{deng2019arcface} features from the images.
        Lastly, both SER-FIQ (Unsupervised Estimation of Face Image Quality Based on Stochastic Embedding Robustness)~\cite{terhorst2020ser} and BRISQUE(Blind/Referenceless Image Spatial Quality Evaluator)~\cite{mittal2012no} measure the quality of images.

    \subsection{Discussion}
\begin{figure*}[t]
            \begin{center}
                \includegraphics[width=9.7cm]{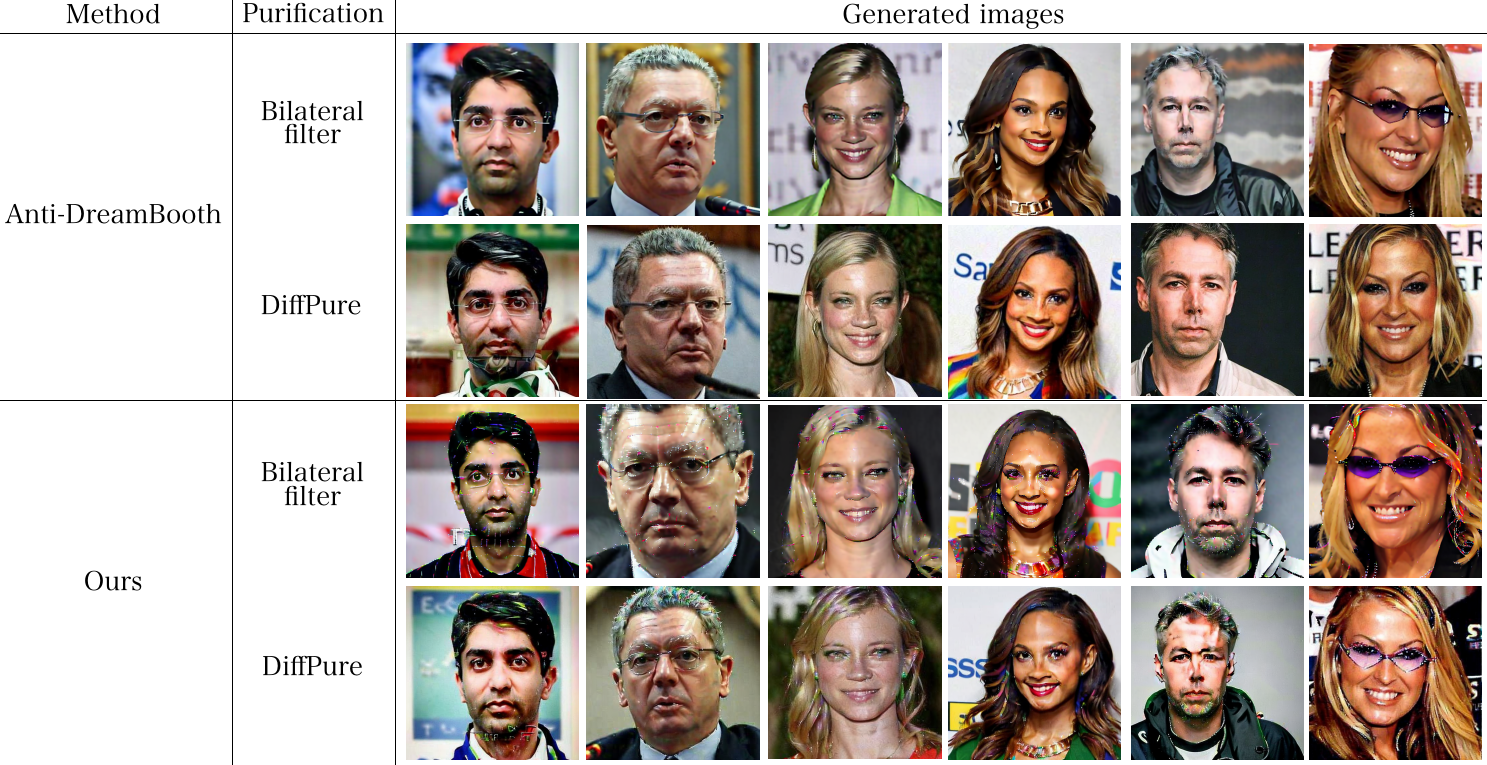}
                \caption{Generated images: Noise budgets were $\eta=0.02$ for Anti-DreamBooth, and $\eta=0.01, \eta_{\rm{mask}}=0.5$ for ours, and the masked area was $3\%$ of images. In this setting, the $L_{1}$ norms of the adversarial examples for the clean images are almost the same.}
                \label{fig:result}
            \end{center}
        \end{figure*}

        Fig. \ref{fig:result} shows the generated images in this experiment.
        Firstly, the first and second rows show that clean images are generated by the model personalized with Anti-DreamBooth's adversarial examples after purifying.
        In other words, the perturbations of adversarial examples are easily removed and lose their defensive capacity.
        On the other hand, in our method, the third and fourth rows show that generated images exhibit unnatural patterns on the faces, indicating successful defense.
        These results suggest that our method is effective in preventing personalized generation.
        Furthermore, our method adds adversarial perturbations that cannot be removed by adversarial purification techniques.

        Tab. \ref{tab:result} is the numerical evaluation of the experiment.
        We couldn't observe a significant difference between those methods. 
        These results suggest that existing metrics cannot assess the authenticity of images, while they can detect significant degradation.
        Thus, considering more sensitive metrics is part of our future work.

        \begin{table}[tb]
            \caption{Comparison between HF-ADB (Ours) and Anti-DreamBooth (ADB). The arrows indicate the direction in which generative quality deteriorates, meaning the defense is successful. The experimental setting is identical to that in Fig. \ref{fig:result}.}
            \begin{adjustbox}{width=0.7\linewidth,center}
            \begin{tabular}{@{}llllll@{}} \toprule    
            Method & Purification & FDSR$\downarrow$ & ISM$\downarrow$ & SER-FIQ$\downarrow$ & BRISQUE$ \uparrow$ \\ \midrule
            ADB & Bilateral filter & $0.97 \pm 0.10$        & $0.35  \pm 0.09$           & $0.76 \pm 0.07$ & $42.3 \pm 7.7$ \\ 
            ADB & DiffPure & $0.96 \pm 0.10$      & $0.36  \pm 0.09$           & $0.76 \pm 0.07$ & $10.7 \pm 4.3$\\ \addlinespace
            Ours & Bilateral filter &$0.96 \pm 0.10$        & $0.35  \pm 0.09$           & $0.76 \pm 0.08$ & $51.5 \pm 8.5$ \\
            Ours & DiffPure & $0.96 \pm 0.10$        & $0.36  \pm 0.09$         & $0.76 \pm 0.08$ & $8.3 \pm 3.7$ \\ \bottomrule
            \end{tabular}
            \label{tab:result}
            \end{adjustbox}
        \end{table}

        Our method has some limitations.
        Firstly, the noise we add is so strong that the noise could be noticeable when viewed up close.
        Secondly, while our method can reduce the effect of adversarial purification, it cannot completely maintain the original defense capacity.

\section{Conclusion}
In this paper, we propose a method to hinder the generation of face images, specifically addressing the problem that image generative models are misused to create fake images.
We found that previous works had an issue that added adversarial perturbation can be removed easily. In contrast, our proposed method achieves a robust adversarial attack against adversarial purification techniques by adding strong adversarial perturbation to the images.
We hope that this method will be widely used and help prevent malicious image generation.

% \clearpage\mbox{}Page \thepage\ of the manuscript.
% \clearpage\mbox{}Page \thepage\ of the manuscript.
% \clearpage\mbox{}Page \thepage\ of the manuscript.
% \clearpage\mbox{}Page \thepage\ of the manuscript.
% \clearpage\mbox{}Page \thepage\ of the manuscript. This is the last page.
% Now we have reached the maximum length of an ECCV \ECCVyear{} submission (excluding references).
% References should start immediately after the main text, but can continue past p.\ 14 if needed.
% \clearpage  % TODO REVIEW/FINAL: This \clearpage needs to be removed from both review and camera-ready versions.

\bibliographystyle{splncs04}
\bibliography{main}

\clearpage
% \section*{Appendix}
% \subsection*{A. Numerical Result}
%     We describe the numerical result of our experiment in this section.
%     Fig. \ref{fig:res_bilateral, fig:res_diffpure} show the experimental results when using a bilateral filter or DiffPure~\cite{nie2022diffusion}, respectively, for adversarial purification.
%     However, these results show no solid difference that determine which method is superior.

% \label{app:result}

% \begin{figure}
%     \centering
%     \includegraphics[width=0.8\linewidth]{img/res_bilateral.pdf}
%     \caption{Evaluation when bilateral filter is applied between our method and Anti-DreamBooth}
%     \label{fig:res_bilateral}
% \end{figure}
% \begin{figure}
%     \centering
%     \includegraphics[width=0.8\linewidth]{img/result_diffpure.pdf}
%     \caption{Evaluation when DiffPure is applied between our method and Anti-DreamBooth}
%     \label{fig:res_diffpure}
% \end{figure}

\setcounter{section}{0}
\renewcommand{\thesection}{\Alph{section}}
\section{Supplemental Material for Defense against Artwork Generation}
    Personalized generation of artworks without permission is also a significant problem as well as fake image generation.
    Here, we applied our method to prevent personalized generation of artworks and analyzed the result.

    \begin{figure*}[t]
        \begin{center}
            \includegraphics[width=9.7cm]{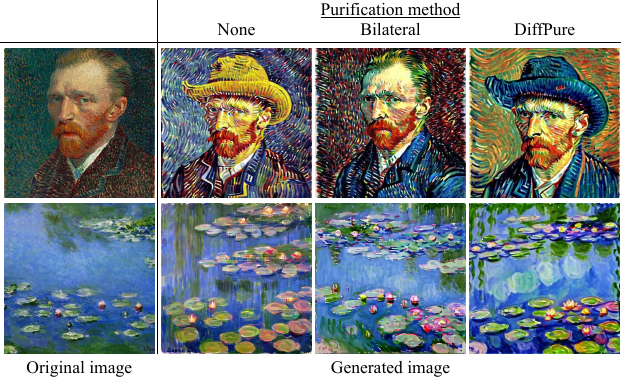}
        \caption{Artworks generated by a model personalized with our method's adversarial examples. Noise budgets were $\eta=0.01$ and $\eta_{\rm{mask}}=0.5$, and the masked area was $3\%$.}
            \label{fig:art}
        \end{center}
    \end{figure*}

Fig. \ref{fig:art} shows the result.
The first column contains the training images used for personalization, and the second to fourth columns show the results generated after applying different purification methods.
We can say that the generated images were relatively clean and the defense failed.
Moreover, the generated images look natural even when we do not apply any purification methods.
Although the generated images have some unnatural patterns, they are not so noisy.
We estimate that the colorful trait of artworks may cause this compared to face images.
Therefore, it may be said that style transformation method like Glaze~\cite{shan2023glaze}, which aims to make personalized models generate different art-style images, is more effective for artwork protection.

\end{document}